\documentclass[letterpaper, 10 pt, journal, twoside]{IEEEtran}
\ifCLASSINFOpdf
\else
\fi

\usepackage{cite}
\usepackage{amsmath,amssymb}
\usepackage{graphicx}
\usepackage{booktabs}
\usepackage[table]{xcolor}
\definecolor{BGM}{rgb}{0.9,0.9,0.9}
\definecolor{darkgreen}{rgb}{0,0.5,0}
\usepackage{multirow,makecell}
\usepackage{url}
\newif\ifdiff
\difffalse                                  
\ifdiff
  \newcommand{\rev}[1]{\textcolor{blue}{#1}}
\else
  \newcommand{\rev}[1]{#1}
\fi
\newif\ifarxiv
\arxivtrue                  

\begin{document}
%
\title{Uni-PrevPredMap: Extending PrevPredMap to a Unified Framework of Prior-Informed Modeling for Online Vectorized HD Map Construction}
%
%
%

\author{Nan Peng$^{1}$, Xun Zhou$^{1,2}$, Mingming Wang$^{3,4}$, Guisong Chen$^{1}$, and Wenqi Xu$^{4}$%
\thanks{Manuscript received: February 2, 2026; Revised June 6, 2026; Accepted August 24, 2026.}%
\thanks{This paper was recommended for publication by Editor Ashis Banerjee upon evaluation of the Associate Editor and Reviewers' comments.}%
\thanks{$^{1}$Nan Peng, Xun Zhou, and Guisong Chen are with Ruqi Mobility, Guangzhou, China
        {\tt\footnotesize pengnan@ruqimobility.com}}%
\thanks{$^{2}$Xun Zhou is also with the School of Intelligent Systems Engineering, Sun Yat-sen University, Shenzhen, China}%
\thanks{$^{3}$Mingming Wang is with the School of Vehicle and Mobility, Tsinghua University, Beijing, China}%
\thanks{$^{4}$Mingming Wang and Wenqi Xu are with GAC Group, Guangzhou, China}%
\ifarxiv
\thanks{\copyright~2026 IEEE. Personal use of this material is permitted. Permission from IEEE must be obtained for all other uses, in any current or future media, including reprinting/republishing this material for advertising or promotional purposes, creating new collective works, for resale or redistribution to servers or lists, or reuse of any copyrighted component of this work in other works.}%
\thanks{This is the accepted version of the article published in IEEE Robotics and Automation Letters, vol.~11, no.~10, pp.~12032--12039, Oct.~2026. Digital Object Identifier: 10.1109/LRA.2026.3729481}
\fi
}
%
%

\ifarxiv
    \markboth{IEEE Robotics and Automation Letters, vol. 11, no. 10, October 2026}
  {Peng \MakeLowercase{\textit{et al.}}: Uni-PrevPredMap}
\fi

%



\maketitle


\begin{abstract}
Safety-critical autonomous driving motivates the effective use of prior information. For online vectorized HD map construction, temporal predictions and cost-efficient HD map priors are two complementary yet individually imperfect sources. However, existing prior-informed approaches typically use only one of them or assume the HD map prior to be reliable. We present Uni-PrevPredMap, a unified framework that treats both as imperfect priors and processes them within a single pipeline through their common vectorized representation. At its core is a tri-mode training paradigm that exposes the model to non-prior, temporal-prior, and temporal-map-fusion conditions. This single design enables one model to perceive reliably without any prior, exploit priors when available, and stay robust when priors are imperfect, rather than being trained under a single fixed prior condition. Uni-PrevPredMap achieves state-of-the-art map-absent performance on nuScenes and Argoverse2. Combining both priors yields gains beyond either source alone, while \rev{robustness to imperfect priors is evaluated under synthetic perturbation types unseen during training}, indicating that the model can benefit from map priors without over-relying on them. Code is available at \url{https://github.com/pnnnnnnn/Uni-PrevPredMap}.
\end{abstract}


%

\section{Introduction}
%
%
%
%
\IEEEPARstart{H}{igh-Definition} (HD) maps serve as critical infrastructure for autonomous vehicles, providing centimeter-level road geometry and semantic information for precise localization and safe navigation. HD maps can be generated either by traditional offline SLAM-based workflows or by emerging online perception systems. Due to the high cost of offline HD map production and maintenance, the automotive industry is increasingly shifting toward online vectorized HD map construction \cite{liao2023maptr}.

\begin{figure}[t]
  \centering
  \includegraphics[height=4cm]{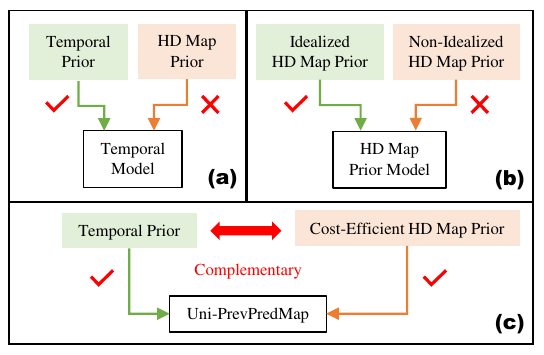}
  \caption{Motivation of Uni-PrevPredMap. (a) Existing temporal models use only the temporal prior. (b) Existing HD map prior models use only the HD map prior and tend to over-rely on idealized prior assumptions, degrading under non-idealized maps. (c) Uni-PrevPredMap jointly leverages temporal predictions and cost-efficient HD map priors, which are complementary yet individually imperfect, within a unified framework.}
  \label{fig:motivation}
\end{figure}

Online vectorized HD map construction degrades under visual deprivation, including severe occlusion and beyond-visual-range conditions. To address this, researchers have proposed two prior sources: temporal perception buffers and cost-efficient HD maps. Here, cost-efficient HD maps refer to lane-level maps obtained through low-cost acquisition pipelines, such as less frequently updated production HD maps and crowd-sourced HD maps, which may exhibit incomplete coverage, partial outdatedness, or local inaccuracies. The two are complementary: temporal buffers ensure backward continuity through sequential observation accumulation, while cost-efficient HD maps provide forward-looking constraints via predefined geometry. However, existing prior-informed models process either temporal buffers \cite{yuan2024streammapnet, wang2024stream, peng2025prevpredmap, chen2024maptracker, zhang2024enhancing, kim2025unveiling} or cost-efficient HD maps \cite{jiang2024p, sun2023mind} in isolation (Fig.~\ref{fig:motivation}(a)). A unified framework jointly leveraging both is therefore desirable for safety-critical autonomy.

Jointly integrating temporal perception buffers with cost-efficient HD map priors presents two challenges: (1) Cost-efficient HD maps deviate from ideal conditions, including geometric inaccuracies and missing or extraneous elements. Current HD map prior models, exemplified in Fig.~\ref{fig:motivation}(b), often rely on accurate and complete priors and may degrade significantly when this assumption is violated \cite{bateman2024exploring}. The unified framework should therefore reduce dependence on idealized prior conditions rather than implicitly trusting the map prior. (2) Cost-efficient HD maps inevitably contain uncovered regions. The unified framework must therefore retain reliable standalone perception where no prior is available, rather than depending on continuous map coverage.

In this work, we present Uni-PrevPredMap, extending the temporal baseline PrevPredMap \cite{peng2025prevpredmap} into a unified prior-informed framework for online vectorized HD map construction. Unlike existing prior-informed approaches that use only one prior source or assume the HD map prior to be reliable, we treat temporal predictions and cost-efficient HD map priors as imperfect yet complementary priors, processing both within a single pipeline through their common vectorized representation. To this end, we introduce a tri-mode training paradigm spanning non-prior, temporal-prior, and temporal-map-fusion conditions under non-idealized priors. The paradigm trains the model to remain robust when priors are imperfect, reflecting the view that a prior-informed model should be trained under multiple prior conditions rather than a single fixed one. The unified pipeline is supported by a tile-indexed 3D global representation, enabling localized retrieval and retaining height information.

In summary, the contributions of this work include:
(1) We present Uni-PrevPredMap, a unified framework that treats temporal predictions and cost-efficient HD map priors as imperfect yet complementary priors for online vectorized HD map construction. Unlike existing prior-informed approaches, which leverage only one of the two sources or assume the HD map prior to be reliable, Uni-PrevPredMap processes both through a common vectorized representation within a single learned pipeline, without explicitly pre-judging their reliability. This representation is organized as a tile-indexed 3D structure to support localized retrieval. 
(2) We introduce a tri-mode training paradigm that exposes the model to non-prior, temporal-prior, and temporal-map-fusion conditions under non-idealized priors. This single training design enables the model to maintain strong standalone perception without any prior, exploit priors when available, and remain robust when they are imperfect.
(3) Extensive experiments show state-of-the-art map-absent performance and synergistic gains from the joint use of both priors beyond either alone (Table~\ref{tab:prior}). \rev{Robustness to imperfect priors is further evaluated under synthetic perturbation types unseen during training} (Table~\ref{tab:simulated}).

%

\section{Related Work}

Online vectorized HD map construction evolved from semantic segmentation \cite{li2022hdmapnet,pan2020cross,li2022bevformer} to end-to-end vectorized prediction by VectorMapNet \cite{liu2023vectormapnet} and one-stage parallel decoding by MapTR \cite{liao2023maptr}. Recent work explores concise map representations \cite{ding2023pivotnet, qiao2023end}, structural query designs \cite{xu2024insmapper}, multi-modal distillation \cite{hao2024mapdistill}, diffusion-based refinement \cite{gao2026refdiffmap}, and segmentation supervision \cite{liao2024maptrv2, choi2024mask2map}.

\subsection{Online Vectorized HD Map Construction with Temporal Perception Buffers}

Temporal perception information is inherently available without additional
acquisition overhead. StreamMapNet \cite{yuan2024streammapnet} performs
dense-sparse co-fusion via streaming integration of BEV and query features. SQD-MapNet \cite{wang2024stream} introduces stream query denoising for temporal consistency. MapTracker \cite{chen2024maptracker} and MapUnveiler \cite{kim2025unveiling} leverage memory buffers to integrate previous BEV and query features. These methods mainly perform temporal fusion at the feature level. PrevPredMap \cite{peng2025prevpredmap} instead adopts a prediction-level formulation, propagating historical outputs as vectorized map elements. As a result, the temporal prior takes the same vectorized form as HD map priors, whereas feature-level representations are not naturally in the same representation space. This representational compatibility allows Uni-PrevPredMap to process temporal predictions and HD map priors through a shared pipeline rather than source-specific pathways.

Map elements are largely static: features or predictions perceived at location $X$ can serve as prior whenever the road structure near $X$ remains unchanged. Building on this, NMP \cite{xiong2023neural} and NeMO \cite{zhu2023nemo} develop region-centric temporal approaches, PreSight \cite{yuan2024presight} introduces NeRF-based city-scale priors, and HRMapNet \cite{zhang2024enhancing} employs a global map processor that stores and distributes rasterized historical predictions. InstGPMap \cite{chen2025instgpmap} extends this to a vectorized form. Uni-PrevPredMap organizes its global representation as a tile-indexed 3D vectorized structure for the unified pipeline, enabling localized retrieval and retaining height information.

\subsection{Online Vectorized HD Map Construction with Cost-efficient Alternative Maps}

Cost-efficient maps, including Standard Definition (SD) maps \cite{luo2024augmenting, jiang2024p}, HD maps \cite{jiang2024p, sun2023mind}, and satellite imagery \cite{gao2024complementing}, provide important priors for online vectorized HD map construction. Among these, cost-efficient HD maps offer the lane-level semantics required for safety-critical autonomy, yet current HD map prior approaches are often developed under the assumption of geometrically accurate, semantically complete, and fully covering prior maps, and may degrade substantially when these assumptions are violated \cite{bateman2024exploring}. One line of work operates on the prior itself: some approaches characterize its reliability, such as uncertainty estimation \cite{monninger2025mapdiffusion} and map change detection \cite{lambert2021trust}, while others maintain and update the prior through probabilistic map maintenance, as in RTMap \cite{du2025rtmap} maintaining a crowdsourced map as a self-evolving memory through change detection and recursive updating. Even so, a maintained prior may remain imperfect at any moment, due to factors such as localization uncertainty, delayed updates, limited spatial coverage, or newly emerging scene changes. This leaves a complementary question that improving the prior alone does not resolve: how should an online predictor behave when the prior it is given is imperfect? Uni-PrevPredMap focuses on this question, treating the prior as an external conditioning signal while preventing the prediction from simply reproducing erroneous prior information. Accordingly, the tri-mode training paradigm spans non-prior, temporal-prior, and temporal-map-fusion conditions, \rev{with robustness evaluated quantitatively under synthetic perturbation types unseen during training (Table~\ref{tab:simulated}) and illustrated qualitatively under an outdated-prior mismatch (Fig.~\ref{fig:real_world})}.

\begin{figure*}[t]
  \centering
  \includegraphics[height=8cm]{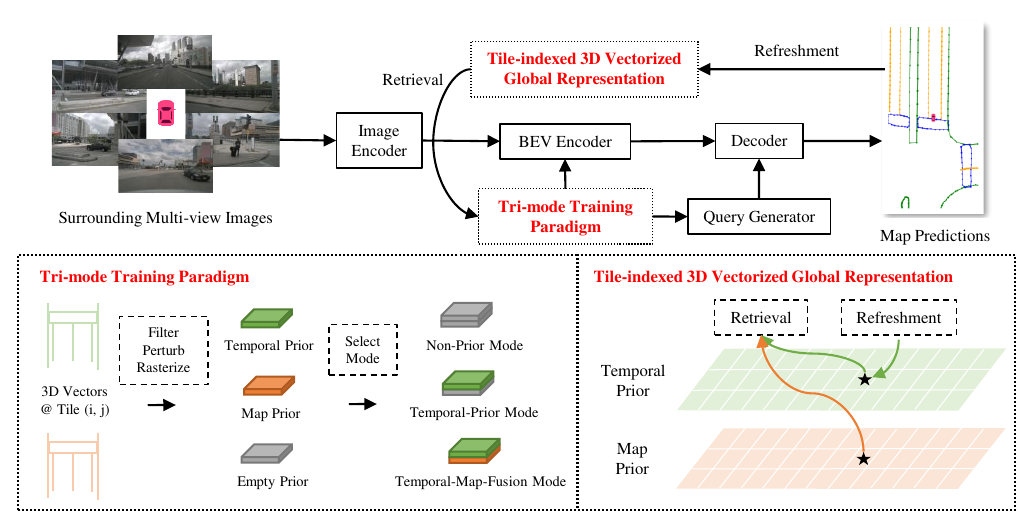}
  \caption{The overall architecture of Uni-PrevPredMap. The lower-left illustrates the tri-mode training paradigm across non-prior, temporal-prior, and temporal-map-fusion conditions. The lower-right highlights the tile-indexed 3D vectorized global representation for localized prior retrieval and refreshment.}
  \label{fig:overview}
\end{figure*}

\section{Method}

\subsection{Overall Architecture}

Uni-PrevPredMap retains PrevPredMap's prediction-level temporal formulation, allowing temporal predictions and cost-efficient HD map priors to share a unified vectorized representation, as illustrated in Fig.~\ref{fig:overview}. A tile-indexed 3D vectorized global representation retrieves both temporal predictions and HD map priors using vehicle positioning coordinates. During training, the tri-mode paradigm exposes the framework to non-prior, temporal-prior, and temporal-map-fusion conditions. The retrieved priors are fused with multi-view image features in the BEV encoder and used to initialize queries, and the decoded predictions are incrementally written back into the global representation.

\subsection{Tri-mode Training Paradigm}

We introduce a tri-mode training paradigm across non-prior, temporal-prior, and temporal-map-fusion conditions. By exposing the framework to heterogeneous prior conditions during training, the paradigm reduces dependence on ideal prior assumptions while supporting robust perception in both map-absent and map-present scenarios.

As illustrated in Fig.~\ref{fig:overview}, retrieved 3D vectors first undergo spatial filtering constrained by the predefined perception range. During training, instance-level perturbation is applied to filtered HD map priors, where random vector subsets receive displacement uniformly sampled from [-6,6] meters. Approximately half of map vectors are perturbed with mean 3-meter displacement, roughly corresponding to standard lane widths. The perturbation process simulates real-world infrastructure modifications including divider adjustments, boundary changes, and pedestrian crossing relocation. During inference, robustness is further evaluated under unseen perturbations including instance-level addition/deletion and frame-level displacement/rotation/scaling operations. Finally, rasterization generates temporal, map, or empty priors for the three training conditions.

According to the selected training condition, temporal, map, and empty priors are combined before concatenation, as illustrated in Fig.~\ref{fig:overview}. During training, Uni-PrevPredMap employs a stratified sampling strategy across three conditions: non-prior, temporal-prior, and temporal-map-fusion.
In the temporal-map-fusion condition, vectors are retrieved from both the temporal buffer and HD map priors within the global representation. These vectors are rasterized into temporal and map heatmaps before being processed by the BEV encoder and query generator. The temporal-prior condition uses only vectors from the temporal buffer to generate the temporal heatmap, while the map heatmap is initialized as zeros. In the non-prior condition, both heatmaps are initialized as zeros.
This tri-mode training paradigm reduces dependence on ideal prior assumptions while supporting robust operation under varying map availability conditions. During inference, the framework utilizes temporal-map-fusion when both temporal and map priors are available, temporal-prior when map priors are absent, and non-prior when neither prior source is available.

The processed temporal and map priors are first concatenated and then incorporated through two parallel pathways: (1) the BEV encoder follows HRMapNet~\cite{zhang2024enhancing} by concatenating the combined prior with BEV features before convolutional fusion; (2) the query generator employs deformable attention, where map queries attend to spatial locations in the combined prior features through learned sampling offsets and attention weights. Neither pathway introduces a hand-crafted weighting between the temporal and map priors. Instead, the convolutional filters in the BEV pathway and the sampling offsets and attention weights in the query pathway are learned jointly with the rest of the network through end-to-end supervision under the tri-mode training paradigm.

\subsection{Tile-indexed 3D Vectorized Global Representation}

Uni-PrevPredMap organizes temporal predictions and HD map priors within a tile-indexed 3D vectorized global representation, supporting localized retrieval and compact storage. Specifically, the representation adopts dual-axis geospatial indexing, where each $(i,j)$-indexed tile corresponds to a bounded geographical region containing discrete map vectors confined to the tile boundaries.

\subsubsection{Refreshment}

High-confidence predictions are incrementally updated into geospatially indexed tiles within the global representation. For temporal predictions, vectors are updated to tiles indexed according to the vehicle's UTM (Universal Transverse Mercator) coordinates. For HD map priors, vectors are pre-stored in the corresponding indexed tiles according to their UTM coordinates. The mapping between tile indices $(i,j)$ and UTM coordinates $(UTM_{\mathrm{east}}, UTM_{\mathrm{north}})$ is defined as:
\begin{equation}
(i, j) = \left( \left\lfloor UTM_{\mathrm{east}} / l \right\rfloor,\ \left\lfloor UTM_{\mathrm{north}} / l \right\rfloor \right),
\label{eq:tile_index}
\end{equation}
where $l$ denotes the long side length of the perception range.

\subsubsection{Retrieval}

Temporal predictions and HD map priors are retrieved from tiles indexed according to the vehicle's UTM coordinates using the refreshment mapping described above. To avoid boundary truncation near tile edges, adjacent tiles surrounding the target tile are additionally retrieved. The retrieval mapping between tile indices $\left\{ (i, j) \mid i \in I,\ j \in J \right\}$ and UTM coordinates $(UTM_{\mathrm{east}}, UTM_{\mathrm{north}})$ is defined as:
\begin{equation}
I = \begin{cases}
(i_t-1,\ i_t) & \mathrm{if}\ UTM_{\mathrm{east}} \bmod l < l/2 \\
(i_t) & \mathrm{if}\ UTM_{\mathrm{east}} \bmod l = l/2 \\
(i_t,\ i_t+1) & \mathrm{if}\ UTM_{\mathrm{east}} \bmod l > l/2
\end{cases}
\end{equation}
\begin{equation}
J = \begin{cases}
(j_t-1,\ j_t) & \mathrm{if}\ UTM_{\mathrm{north}} \bmod l < l/2 \\
(j_t) & \mathrm{if}\ UTM_{\mathrm{north}} \bmod l = l/2 \\
(j_t,\ j_t+1) & \mathrm{if}\ UTM_{\mathrm{north}} \bmod l > l/2
\end{cases}
\end{equation}
where $i_t$ and $j_t$ denote the indices of the target tile.

\section{Experiment}

\subsection{Experimental Setup}

\subsubsection{Datasets}
We evaluate Uni-PrevPredMap on two large-scale benchmarks: nuScenes~\cite{caesar2020nuscenes} and Argoverse2~\cite{wilson2023argoverse}. The nuScenes dataset provides 2D vectorized maps and contains 1000 scenes, with 700 used for training and 150 for validation. Each scene consists of 20 seconds of 2Hz RGB images captured by 6 cameras. Argoverse2 provides 3D vectorized maps and contains 1000 logs, with 700 used for training and 150 for validation. Each log consists of 15 seconds of 20Hz RGB images captured by 7 ring cameras.

\subsubsection{Evaluation Metrics}
Consistent with previous methods~\cite{li2022hdmapnet,liu2023vectormapnet,liao2023maptr}, we evaluate three static map categories: pedestrian crossings, lane dividers, and road boundaries. The perception range is set to 30m front/rear and 15m left/right of the vehicle. Average precision (AP) based on Chamfer Distance is adopted as the evaluation metric under three thresholds of $\{0.5, 1.0, 1.5\}$m.

\subsubsection{Implementation Details}
We adopt ResNet50~\cite{he2016deep} as the image backbone and LSS-based BEVPoolv2~\cite{huang2022bevpoolv2} for the PV-to-BEV transformation. AdamW with weight decay 0.01 is used as the optimizer, with an initial learning rate of 0.0006 and a cosine decay schedule. The batch size is set to 16, and all models are trained on 4 NVIDIA A100 GPUs. The BEV grid size is set to 0.3m. The numbers of instance queries, point queries, and decoder layers are set to 100, 20, and 6, respectively.

\begin{table}[t]
  \caption{Comparison with SOTA methods on nuScenes. All methods use ResNet50 backbone, with FPS measured on an NVIDIA RTX A6000. Uni-PrevPredMap and Uni-PrevPredMap$^{*}$ report map-absent and map-present inference results, respectively, from the same unified model. $\star$ taken from corresponding papers, FPS scaled based on MapTRv2 \cite{liao2024maptrv2}.}
  \label{tab:nuscenes}
  \centering
  \setlength{\tabcolsep}{4pt} 
  \begin{tabular}{ l c c c c c c c c }
    \toprule
    Method & Epoch & $AP_{div}$ & $AP_{ped}$ & $AP_{bou}$ & mAP & FPS \\
    \midrule
    MapTRv2 \cite{liao2024maptrv2} & 24 & 62.4 & 59.8 & 62.4 & 61.3 & 16.4 \\
    HRMapNet \cite{zhang2024enhancing} & 24 & 67.4 & 65.8 & 68.5 & 67.2 & 13.3 \\
    Mask2Map \cite{choi2024mask2map} & 24 & 71.3 & 70.6 & 72.9 & 71.6 & 9.5 \\
    MapTracker \cite{chen2024maptracker} & 24 & 69.2 & 75.3 & 71.2 & 71.9 & 11.7 \\
    MapUnveiler \cite{kim2025unveiling} & 24 & 67.6 & 67.6 & 68.8 & 68.0 & 13.4$^\star$ \\
    PriorMapNet \cite{wang2024priormapnet} & 24 & 69.0 & 64.0 & 68.2 & 67.1 & 13.7$^\star$ \\
    FastMap \cite{hu2025fastmap} & 24 & 69.1 & 65.5 & 69.7 & 68.1 & 17.2$^\star$ \\
    \rowcolor{BGM}
    Uni-PrevPredMap & 24 & {\bf 72.3} & {\bf 76.2} & {\bf 73.6} & {\bf 74.0} & 12.2 \\
    \rowcolor{BGM}
    Uni-PrevPredMap* & 24 & 84.1 & 79.1 & 79.6 & 80.9 & 11.5 \\
    \midrule
    MapTRv2 \cite{liao2024maptrv2} & 110 & 68.8 & 68.0 & 71.0 & 69.2 & 16.4 \\
    HIMap \cite{zhou2024himap} & 110 & 75.0 & 71.3 & 74.7 & 73.7 & 10.0$^\star$ \\
    HRMapNet \cite{zhang2024enhancing} & 110 & 72.9 & 72.0 & 75.8 & 73.6 & 13.3 \\
    Mask2Map \cite{choi2024mask2map} & 110 & 73.6 & 73.1 & {\bf 77.3} & 74.6 & 9.5 \\
    MapTracker \cite{chen2024maptracker} & 72 & 74.1 & {\bf 80.0} & 74.1 & 76.1 & 11.7 \\
    MGMapNet \cite{yang2024mgmapnet} & 110 & 74.3 & 71.8 & 74.8 & 73.6 & 13.3$^\star$ \\
    \rowcolor{BGM}
    Uni-PrevPredMap & 72 & {\bf 76.3} & 77.9 & 76.6 & {\bf 77.0} & 12.2 \\
    \rowcolor{BGM}
    Uni-PrevPredMap* & 72 & 83.4 & 79.7 & 82.3 & 81.8 & 11.5 \\
    \bottomrule
  \end{tabular}
\end{table}

\subsection{Comparisons with State-of-the-art Methods}

\subsubsection{Performance on nuScenes}

As shown in Table~\ref{tab:nuscenes}, Uni-PrevPredMap achieves 74.0 mAP (24 epochs) and 77.0 mAP (72 epochs), surpassing existing methods in training convergence, validation accuracy, and inference speed. When integrated with non-idealized HD map priors, the enhanced variant Uni-PrevPredMap* further achieves 80.9 mAP (24 epochs) and 81.8 mAP (72 epochs), \rev{indicating that the model benefits from map priors even when they are synthetically perturbed}.
While Uni-PrevPredMap achieves strong overall performance, it does not outperform every prior method across all categories. Under the map-absent setting, the Boundary AP is 0.7 lower than Mask2Map~\cite{choi2024mask2map}. We attribute this primarily to the difference in prior conditioning strategies. Mask2Map leverages current-frame segmentation features as priors, providing highly precise and up-to-date spatial cues for boundary modeling, whose geometry often varies significantly across frames. In contrast, Uni-PrevPredMap relies on historical temporal priors, which provide comparatively weaker spatial guidance in map-absent regions. However, when non-idealized HD map priors are available, Uni-PrevPredMap surpasses Mask2Map by 5.0 AP, indicating that even imperfect map priors contribute structural information not captured by temporal predictions.
For pedestrian crossings, Uni-PrevPredMap underperforms MapTracker~\cite{chen2024maptracker} by 2.1 AP under the map-absent setting. We attribute this to the different temporal representations adopted by the two methods. MapTracker propagates query-level features that preserve vector-level semantics during temporal modeling, which is advantageous for pedestrian crossings whose box-style representations are more difficult to reconstruct from rasterized priors. In contrast, Uni-PrevPredMap propagates prediction-level temporal priors that are rasterized before feature extraction, making the reconstruction of fine-grained crossing geometry more challenging. When non-idealized HD map priors are introduced, the gap narrows to 0.3 AP.

\begin{table}[t]
  \caption{Comparison with SOTA methods on Argoverse2. All backbones utilized are ResNet50. }
  \label{tab:argoverse2}
  \centering
  \begin{tabular}{ l c c c c }
    \toprule
    Method & $AP_{div}$ & $AP_{ped}$ & $AP_{bou}$ & mAP \\
    \midrule
    MapTRv2 \cite{liao2024maptrv2} & 68.9 & 60.7 & 64.5 & 64.7 \\
    HIMap \cite{zhou2024himap} & 68.3 & 66.7 & 70.3 & 68.4 \\
    MapUnveiler \cite{kim2025unveiling} & 72.6 & 66.0 & 67.6 & 68.7 \\
    MGMapNet \cite{yang2024mgmapnet} & 72.1 & 64.7 & 70.4 & 69.1 \\
    PriorMapNet \cite{wang2024priormapnet} & 73.4 & 66.5 & 69.8 & 69.9 \\
    \rowcolor{BGM}
    Uni-PrevPredMap & {\bf 74.5} & {\bf 69.5} & {\bf 72.9} & {\bf 72.3} \\
    \rowcolor{BGM}
    Uni-PrevPredMap* & 82.5 & 75.2 & 80.5 & 79.4 \\
    \bottomrule
  \end{tabular}
\end{table}

\subsubsection{Performance on Argoverse2}

Argoverse2 provides 3D vectorized maps with elevation information. As shown in Table~\ref{tab:argoverse2}, Uni-PrevPredMap achieves strong performance under the 3D evaluation protocol. Following MapTRv2, Uni-PrevPredMap is trained for 6 epochs and evaluated at a 2.5Hz sampling rate. Uni-PrevPredMap achieves 72.3 mAP, outperforming all existing methods on the Argoverse2 benchmark. When integrated with non-idealized HD map priors, Uni-PrevPredMap* further improves performance to 79.4 mAP.

\begin{table}[tb]
  \caption{Framework evolution: PrevPredMap to Uni-PrevPredMap.}
  \label{tab:roadmap}
  \centering
  \begin{tabular}{ l c c }
    \toprule
    Setting & mAP (w/o map) & mAP (w/ map) \\
    \midrule
    PrevPredMap & 66.3 & - \\
    \midrule
    + tile-indexed 3D processor & 74.0 & - \\
    + tri-mode paradigm & 74.0 & 80.9 \\
    \midrule
    = Uni-PrevPredMap & \textcolor{darkgreen}{+7.7} & \textcolor{darkgreen}{n/a $\rightarrow$ 80.9} \\
    \bottomrule
  \end{tabular}
\end{table}

\subsection{Ablation Study}

\subsubsection{Framework Evolution: PrevPredMap to Uni-PrevPredMap}

Starting from PrevPredMap~\cite{peng2025prevpredmap}, we incrementally introduce the tile-indexed 3D vectorized global representation and the tri-mode training paradigm, with results summarized in Table~\ref{tab:roadmap}. The global representation enables historical predictions to be incorporated through both the BEV encoder and query generator, extending beyond PrevPredMap's single-frame query attention mechanism. This improves map-absent performance from 66.3 to 74.0 mAP.
The tri-mode training paradigm further enables robust utilization of HD map priors, improving performance to 80.9 mAP when map priors are available while preserving strong performance (74.0 mAP) in the map-absent setting. These results indicate that Uni-PrevPredMap benefits from map priors without sacrificing standalone perception capability.

\begin{table}[t]
  \caption{Comparison across tri-mode paradigm variants.}
  \label{tab:tri-mode}
  \centering
  \begin{tabular}{c c c}
    \toprule
    Training Paradigm & mAP (w/o map) & mAP (w/ map) \\
    \midrule
    Single-mode & 16.6 & 82.9 \\
    Dual-mode-a & 60.6 & 77.4 \\
    Dual-mode-b & 69.0 & 84.4 \\
    \rowcolor{BGM}
    Tri-mode & 74.0 & 80.9 \\
    \bottomrule
  \end{tabular}
\end{table}

\subsubsection{Comparative Performance across Tri-mode Paradigm Variants} 

Table~\ref{tab:tri-mode} evaluates degraded variants of the proposed tri-mode training paradigm. The single-mode variant is trained only under map-prior conditions, while the dual-mode variants include dual-mode-a (non-prior and map-prior conditions) and dual-mode-b (temporal-prior and temporal-map-fusion conditions).
As shown, the single-mode variant, representative of many existing HD map prior approaches~\cite{jiang2024p,sun2023mind}, performs poorly in the map-absent setting. This suggests that relying predominantly on map priors during training can reduce robustness when prior information becomes unavailable or deviates from idealized conditions.
The performance gap of dual-mode-a further highlights the importance of temporal priors, indicating that temporal and map priors are not mutually redundant within the unified framework. Meanwhile, the relatively smaller gap of dual-mode-b suggests that explicit exposure to non-prior conditions during training remains beneficial even when temporal priors are available, at the cost of a moderate reduction in map-present accuracy.
Overall, the tri-mode training paradigm reduces dependence on ideal prior assumptions while maintaining strong performance in both map-absent and map-present settings.

\begin{table}[t]
  \caption{Inference performance under different prior conditions.}
  \label{tab:prior}
  \centering
  \begin{tabular}{c c c c}
    \toprule
    \shortstack{Temporal Prior} & \shortstack{Map Prior} & mAP & FPS \\
    \midrule
    \textcolor{red}{$\times$} & \textcolor{red}{$\times$} & 64.9 & 14.2 \\
    \textcolor{darkgreen}{\checkmark} & \textcolor{red}{$\times$} & 74.0 & 12.2 \\
    \textcolor{red}{$\times$} & \textcolor{darkgreen}{\checkmark} & 71.3 & 13.1 \\
    \textcolor{darkgreen}{\checkmark} & \textcolor{darkgreen}{\checkmark} & 80.9 & 11.5 \\
    \bottomrule
  \end{tabular}
\end{table}

\begin{table}[t]
  \caption{Robustness to synthetic HD map perturbations.}
  \label{tab:simulated}
  \centering
  \begin{tabular}{c c c}
    \toprule
    Type & Noise Range & mAP \\
    \midrule
    \rowcolor{BGM}
    Perfect Map & - & 85.6 \\ \midrule
    \multirow{3}{*}{Instance-Level Displacement} & [-3m,3m] & 81.9 \\ \cmidrule{2-3}
    & [-6m,6m] & 80.9 \\ \cmidrule{2-3}
    & [-9m,9m] & 80.3 \\ \midrule
    Instance-Level Addition & [0,10] instances & 85.2 \\ \midrule
    Instance-Level Deletion & [0,10] instances & 82.5 \\ \midrule
    \multirow{3}{*}{Frame-Level Displacement} & [-3m,3m] & 77.0 \\ \cmidrule{2-3}
    & [-6m,6m] & 75.7 \\ \cmidrule{2-3}
    & [-9m,9m] & 74.6 \\ \midrule
    Frame-Level Rotation & [-15$^\circ$,15$^\circ$] & 77.0 \\ \midrule
    Frame-Level Scale & [0.8,1.2] & 77.0 \\ \midrule
    \rowcolor{BGM}
    Empty Map & - & 74.0 \\
    \bottomrule
  \end{tabular}
\end{table}

\subsubsection{Inference Performance under Different Prior Conditions} 

Table~\ref{tab:prior} shows that the tri-mode training paradigm maintains strong performance in both map-absent (74.0 mAP) and map-present (80.9 mAP) settings. Combining temporal and map priors (80.9 mAP) outperforms each used individually (temporal: 74.0 mAP; map: 71.3 mAP), improving by 6.9 mAP over the stronger individual prior. This gain empirically confirms the complementarity motivated in Section~I.

\subsubsection{Robustness to Synthetic HD Map Perturbations}

Uni-PrevPredMap is trained with instance-level displacement uniformly sampled from $[-6,6]$m, while additional unseen perturbations are introduced during inference to evaluate robustness under non-idealized HD map priors. Specifically, we consider instance-level addition/deletion and frame-level displacement/rotation/scaling operations. These perturbation types are designed to approximate representative categories of real-world map inconsistencies, including localized geometric changes, topological inconsistencies, and pose-related systematic biases.
As shown in Table~\ref{tab:simulated}, Uni-PrevPredMap achieves 85.6 mAP even with perfect HD map priors, indicating that the framework continues to rely on perception cues rather than solely depending on map information. Performance varies naturally with perturbation magnitude, where smaller displacements cause limited degradation while larger deviations lead to more noticeable performance drops.
Instance-level addition and deletion simulate newly constructed and removed road elements that alter map connectivity, corresponding to topological inconsistencies. Uni-PrevPredMap remains robust under both perturbations, with only 0.4 mAP degradation under additions and 3.1 mAP degradation under deletions. Frame-level displacement, rotation, and scaling simulate systematic map-to-vehicle misalignment affecting all map elements consistently. Under these perturbations, Uni-PrevPredMap continues to outperform the map-absent baseline, indicating that it exploits informative priors when available while remaining robust to erroneous map information.

\begin{figure}[t]
  \centering
  \includegraphics[height=4.3cm]{0aa4e8f5.pdf}
  \caption{Qualitative robustness under different prior conditions on Argoverse2. 
    The top row (\textit{Map Prior}) shows the input prior, and the bottom row (\textit{Uni-PrevPredMap Prediction}) shows the corresponding clip-level prediction aggregated at a confidence threshold of 0.6. 
    From left to right, the prior conditions are up-to-date, mirrored outdated, and empty. 
    Clip-wise mirroring produces a structurally self-consistent but strongly inconsistent prior; under this prior, the prediction is not rewritten to follow the corrupted geometry but remains aligned with the true road layout, behaving closer to the empty-prior case while preserving topological consistency.}
  \label{fig:real_world}
\end{figure}

\subsubsection{Qualitative Robustness via an Outdated-Prior Stress Test}

Since no public benchmark provides genuine outdated maps paired with sensor data, we construct a surrogate outdated-prior stress test via clip-wise mirroring. Specifically, outdated priors are obtained by reflecting a clip's annotated ground-truth map about the trajectory starting point along the initial driving direction. Unlike the local perturbations in Table~\ref{tab:simulated}, this operation yields a structurally self-consistent, road-like prior that is strongly inconsistent with the current scene, serving as a stress test of whether the model is misled by a plausible but incorrect prior. Fig.~\ref{fig:real_world} shows a representative example under up-to-date, mirrored outdated, and empty prior conditions. Under the mirrored prior, the predicted topology is not rewritten to follow the corrupted geometry, such as the incorrect road alignment or intersection configuration, but remains aligned with the true road layout. Its behavior is closer to the empty-prior case, suggesting that under a strongly inconsistent prior the model relies more on the online perception stream than on the corrupted prior. Aggregations under the mirrored and empty priors are somewhat noisier than under the up-to-date prior but preserve topological consistency. Additional cases are provided in the supplementary materials.

\begin{table}[t]
  \caption{Ablation on the tri-mode sampling ratio with T:T\&M fixed at 3:2 on nuScenes. ``N'', ``T'', and ``M'' denote non-prior, temporal prior, and map prior respectively.}
  \label{tab:probability1}
  \centering
  \begin{tabular}{ c c c c }
    \toprule
      Ratio (N:T:T\&M) & mAP (T) & mAP (M) & mAP (T\&M) \\
    \midrule
    0.30 : 0.42 : 0.28 & 73.4 & 72.4 & 82.7 \\
    0.40 : 0.36 : 0.24 & 73.7 & 71.8 & 82.0 \\
    0.50 : 0.30 : 0.20 & 74.0 & 71.3 & 80.9 \\
    0.60 : 0.24 : 0.16 & 73.9 & 70.3 & 79.7 \\
    0.70 : 0.18 : 0.12 & 72.7 & 69.4 & 78.0 \\
    \bottomrule
  \end{tabular}
\end{table}

\begin{table}[t]
  \caption{Ablation on the tri-mode sampling ratio with N:(T+T\&M) fixed at 1:1. ``T'', ``M'', and ``T\&M'' denote temporal-prior, map-prior, and temporal-map-fusion inference modes.}
  \label{tab:probability2}
  \centering
  \setlength{\tabcolsep}{3.5pt}
  \begin{tabular}{c c c c c c c}
    \toprule
    \multirow{2}{*}{Ratio (N:T:T\&M)} & \multicolumn{3}{c}{nuScenes} & \multicolumn{3}{c}{Argoverse2} \\
    \cmidrule(lr){2-4} \cmidrule(lr){5-7}
     & T & M & T\&M & T & M & T\&M \\
    \midrule
    0.50 : 0.35 : 0.15 & 73.1 & 69.8 & 79.4 & 71.5 & 72.9 & 77.8 \\
    0.50 : 0.30 : 0.20 & 74.0 & 71.3 & 80.9 & 72.4 & 73.6 & 78.6 \\
    0.50 : 0.25 : 0.25 & 73.6 & 72.0 & 81.4 & 72.3 & 74.7 & 79.4 \\
    \rowcolor{BGM}
    0.50 : 0.50 : 0.00 & 74.0 & - & - & 71.9 & - & - \\
    \bottomrule
  \end{tabular}
\end{table}

\begin{figure*}[t]
  \centering
  \includegraphics[height=10cm]{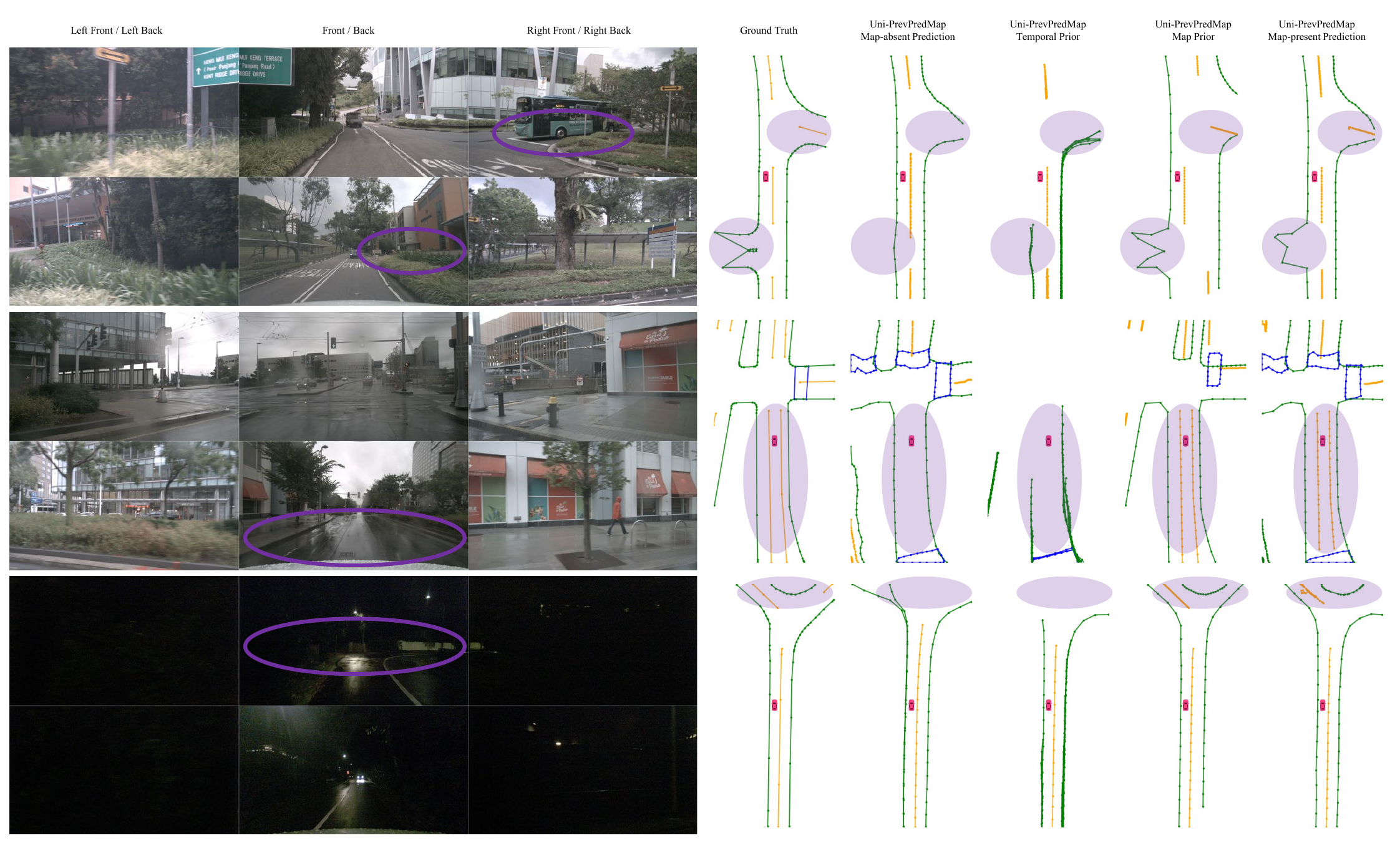}
  \caption{Comparison of predictions by Uni-PrevPredMap with versus without non-idealized HD maps, with differences highlighted in purple shading across occlusion, rainy, and night scenarios. Corresponding priors are illustrated to demonstrate their influence. Green, orange, and blue lines represent road boundaries, lane dividers, and pedestrian crossings, respectively. }
  \label{fig:qa}
\end{figure*}

\subsubsection{Ablation on the Sampling Ratio of the Tri-mode Training Paradigm}

During training, Uni-PrevPredMap employs stratified sampling across non-prior, temporal-prior, and temporal-map-fusion conditions. We conduct ablations from two perspectives.
First, with the ratio between temporal-prior (T) and temporal-map-fusion (T\&M) conditions fixed at 3:2, we vary the proportion between the non-prior condition (N) and the combined prior-based conditions (T+T\&M). As shown in Table~\ref{tab:probability1}, the best map-absent performance is achieved at N:(T+T\&M) = 0.5:0.5. Reducing the proportion of (T+T\&M) causes the model to overfit the non-prior condition, degrading performance under temporal-prior and temporal-map-fusion settings. Conversely, increasing (T+T\&M) leads to stronger dependence on priors: while temporal-map-fusion performance improves slightly, temporal-prior performance declines. We attribute this to the fact that temporal-prior inference still depends on an initial non-prior prediction, making standalone perception capability important for subsequent temporal propagation.
Second, with N:(T+T\&M) fixed at 1:1, we vary the T:T\&M ratio on both nuScenes and Argoverse2 (Table~\ref{tab:probability2}). Consistent trends are observed across both datasets. The best map-absent performance is achieved at N:T:T\&M = 0.5:0.3:0.2. Increasing T\&M leads to stronger dependence on map priors, slightly degrading temporal-prior performance. Conversely, reducing T\&M weakens the supervisory effect provided by map priors and also degrades performance. When T\&M is completely removed, the framework degenerates into a temporal-only configuration.

\subsubsection{Sensitivity to the Refreshment Threshold}

The temporal-prior refreshment uses a confidence threshold of 0.4 during training and 0.9 during inference. The training threshold influences the final result, with 0.4 being empirically optimal. As shown in Table~\ref{tab:threshold}, different training thresholds correspond to different favorable inference thresholds, indicating non-trivial coupling between training-time refreshment behavior and the inference-time update policy. Once the training threshold is fixed at 0.4, performance is stable across the tested range of inference thresholds, varying by no more than 0.3 in both temporal-prior (T) and temporal-map-fusion (T\&M) modes, indicating that the refreshment mechanism is not narrowly tuned. This simple single-threshold refreshment is intentional: rather than seeking a perfectly refreshed temporal prior, the framework relies on the tri-mode paradigm to tolerate imperfect priors, so that robustness arises from this tolerance rather than from precise threshold tuning.

\begin{table}[t]
  \caption{Sensitivity to the refreshment threshold on nuScenes. ``T'' and ``T\&M'' denote temporal-prior and temporal-map-fusion modes respectively.}
  \label{tab:threshold}
  \centering
  \begin{tabular}{c c c c c c c}
    \toprule
    \multicolumn{2}{c}{Training 0.3} & \multicolumn{2}{c}{Training 0.4} & \multicolumn{2}{c}{Training 0.5} \\
    \cmidrule(lr){1-2} \cmidrule(lr){3-4} \cmidrule(lr){5-6}
    Inf. & T / T\&M & Inf. & T / T\&M & Inf. & T / T\&M \\
    \midrule
    0.5 & 72.9 / 80.3 & 0.7 & 73.7 / 81.0 & 0.5 & 72.6 / 79.2 \\
    0.6 & 73.0 / 80.1 & 0.8 & 74.0 / 81.0 & 0.6 & 73.4 / 79.5 \\
    0.7 & 72.8 / 80.0 & 0.9 & 74.0 / 80.9 & 0.7 & 73.2 / 79.6 \\
    \bottomrule
  \end{tabular}
\end{table}

\subsubsection{Qualitative Analysis}

Fig.~\ref{fig:qa} visualizes predictions and priors from Uni-PrevPredMap under occlusion, rainy, and night conditions, comparing results with and without non-idealized HD map priors. Purple shading highlights the differences. Without map priors, prediction quality degrades noticeably in these challenging environments. Incorporating non-idealized HD map priors provides additional structural cues, resulting in clearer and more complete predictions. Additional qualitative examples are provided in the supplementary materials.

\section{Conclusion}
This paper presents Uni-PrevPredMap, a unified prior-informed framework for online vectorized HD map construction. Uni-PrevPredMap treats temporal predictions and cost-efficient HD map priors as complementary yet imperfect sources, and processes them within a single pipeline through their common vectorized representation. At its core, a tri-mode training paradigm exposes the model to non-prior, temporal-prior, and temporal-map-fusion conditions, enabling a single model to perceive reliably without any prior, exploit priors when available, and remain robust when priors are imperfect. The unified pipeline is supported by a tile-indexed 3D global representation for localized retrieval.
Extensive experiments demonstrate state-of-the-art map-absent performance on nuScenes and Argoverse2. When non-idealized HD map priors are available, Uni-PrevPredMap further improves performance, indicating that the model can benefit from map priors without over-relying on them. \rev{A full quantitative evaluation under genuinely evolving real-world maps is beyond the scope of this letter and remains future work; the outdated-prior behavior reported here is a qualitative stress test on a synthetic surrogate rather than a claim of validated real-world robustness. A current limitation is that public benchmarks do not provide longitudinal sensor-map pairs with naturally evolving historical maps, which we hope future benchmark efforts will address.} We hope this work provides useful insights for future autonomous driving perception systems operating under non-idealized prior conditions.

\bibliographystyle{IEEEtran}
\bibliography{references}

%








\end{document}